\documentclass[11pt]{article}       
\usepackage{geometry}               
\usepackage{cite}
\usepackage{amsmath,amssymb,amsfonts}
\usepackage{amsthm}
\usepackage{algorithmic}
\usepackage{graphicx}
\usepackage{textcomp}
\usepackage[table,xcdraw]{xcolor}
\usepackage{tikz}
\usepackage{subcaption}
\usepackage{mathtools}
\usepackage{multirow}
\usepackage{multicol}
\usepackage{marginnote,zref-savepos}
\usepackage{tabu}

\usepackage[table,xcdraw]{xcolor}

\newtheorem{lemma}{Lemma}
\newtheorem{assumption}{Assumption}
\newtheorem{proposition}{Proposition}
\newtheorem{definition}{Definition}

\newcommand{\dist}{\mathrm{dist}}
\newcommand{\st}{\mathrm{s.t.}}

\newtheorem{remark}{Remark}

\title{Risk-Aware Wasserstein Distributionally Robust \\ Control of Vessels in Natural Waterways}

\author{J.M. Nadales, A. Hakobyan, D. Mu\~{n}oz de la Pe\~{n}a, D. Limon, and I. Yang\thanks{The work of J.M. Nadales, D. Mu\~{n}oz de la Pe\~{n}a, and D. Limon was supported by Fondo Europeo de Desarrollo Regional FEDER under Programa Interreg V-A Esp\~{a}na-Portugal
(POCTEP) 2014-2020. The work of A. Hakobyan and I. Yang was supported in part by the National Research Foundation of Korea under Grant MSIT2020R1C1C1009766, in part by the Information and Communications Technology Planning and Evaluation under Grant MSIT2022-0-00480, and in part by Samsung Electronics. \\ 
J.M. Nadales, D. Limon, and D. Mu\~{n}oz de la Pe\~{n}a  are with the Department of Automatic Control and System Engineering, University of Seville. A. Hakobyan and I. Yang are with the Department of Electrical and Computer Engineering at Seoul National University.
Emails: {\tt\small nadales@us.es, dmunoz@us.es, dlm@us.es, astghikhakobyan@snu.ac.kr, insoonyang@snu.ac.kr}.}
}

\date{}

\begin{document}
\maketitle

\pagestyle{myheadings}
\thispagestyle{plain}

\begin{abstract}
In the realm of maritime transportation, autonomous vessel navigation in natural inland waterways faces persistent challenges due to unpredictable natural factors. Existing scheduling algorithms fall short in handling these uncertainties, compromising both safety and efficiency. Moreover, these algorithms are primarily designed for non-autonomous vessels, leading to labor-intensive operations vulnerable to human error.
To address these issues, this study proposes a risk-aware motion control approach for vessels that accounts for the dynamic and uncertain nature of tide islands in a distributionally robust manner. Specifically, a model predictive control method is employed to follow the reference trajectory in the time-space map while incorporating a risk constraint to prevent grounding accidents. To address uncertainties in tide islands, a novel modeling technique represents them as stochastic polytopes. Additionally, potential inaccuracies in waterway depth are addressed through a risk constraint that considers the worst-case uncertainty distribution within a Wasserstein ambiguity set around the empirical distribution.
Using sensor data collected in the Guadalquivir River, we empirically demonstrate the performance of the proposed method through simulations on a vessel. As a result, the vessel successfully navigates the waterway while avoiding grounding accidents, even with a limited dataset of observations. This stands in contrast to existing non-robust controllers, highlighting the robustness and practical applicability of the proposed approach.
\end{abstract}

\section{Introduction}\label{sec:intro}

\subsection{Challenges in Navigating Tidal Inland Waterways}
In today's interconnected and globalized society, maritime transportation and international logistics are vital for global economic development~\cite{park2019role}. In the case of inland ports connected to the sea by artificial channels or natural waterways, traffic management is the crucial factor for port efficiency. This issue has gained significant attention due to the substantial impact of transit channels on the global economy~\cite{li2022impact}, concerns about the potential adverse impact of maritime transportation on climate change~\cite{christodoulou2020forecasting}, and strong interest of maritime authorities in risk management processes to prevent shipping accidents~\cite{kulkarni2020preventing}.

 \begin{figure}[t]
     \centering
 	\includegraphics[width=0.6\linewidth]{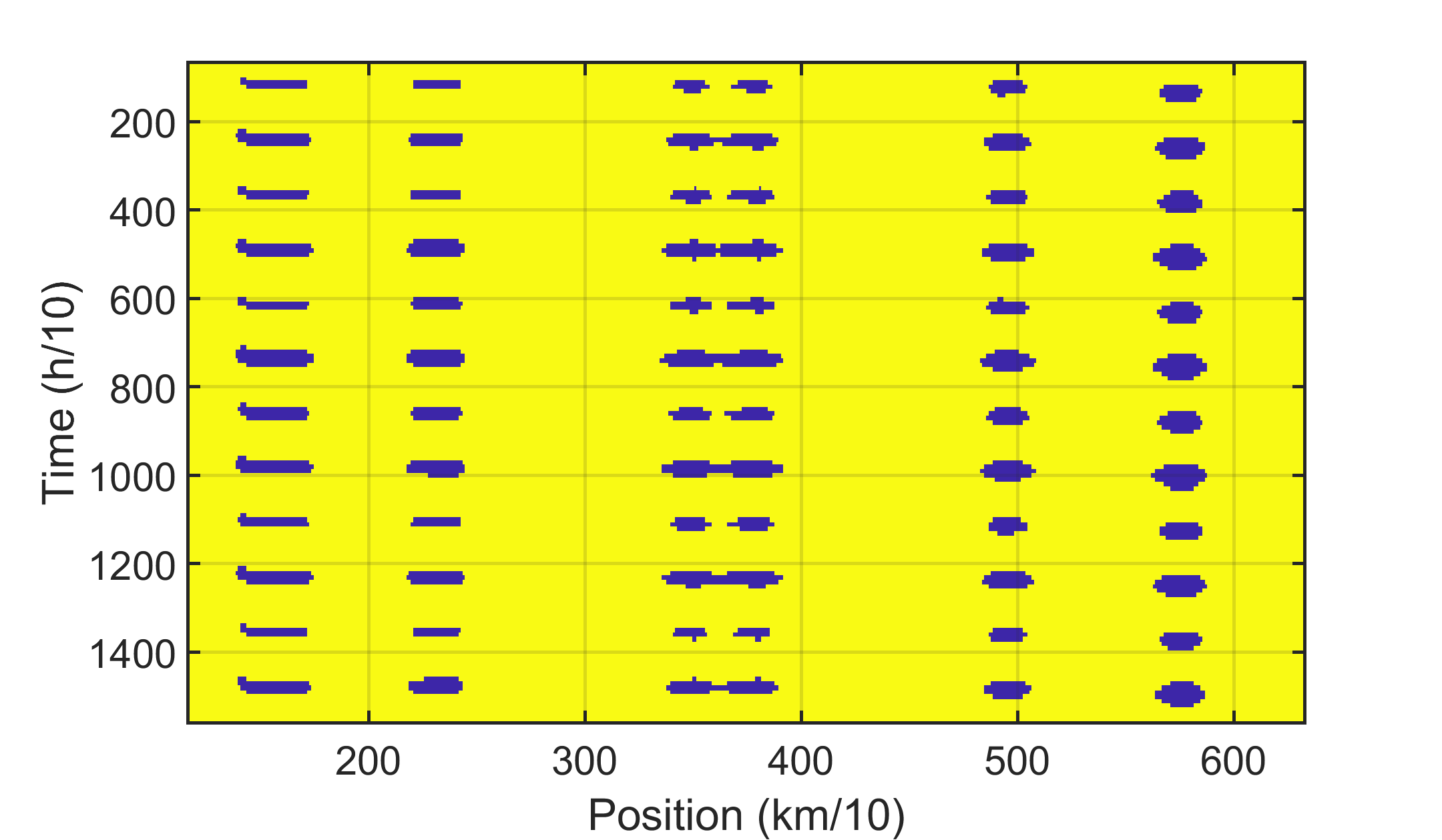}
 	\caption{Depth map example for a 7-meter draft vessel.}
 	\label{fig:islands}
\end{figure}

As a striking example, in this article, we consider the Guadalquivir River in the South of Spain, which is characterized by a changing depth. In this river, the spatial and temporal variation of the depth is mainly due to two natural phenomena: the irregular bathymetric profile of the waterway~\cite{costa2009establecimiento} and the effect of the tide~\cite{losada2017tidal}. 
Due to its location in a highly protected area, the Do\~{n}ana National Park, dredging operations~\cite{donazar2018maintenance} are not allowed, making the problem even more challenging. Additionally, due to the risk of grounding accidents caused by the varying depth of the river, expert river pilots are required to control the vessels in the waterway, and thus, navigation in this channel is prone to human errors.

In order to prevent grounding accidents, it is essential to create a map of time-space points, known as \emph{tide islands}, which takes into account the depth at each location of the waterway and the draft of each vessel at every time instant. This map is produced by overlaying the depth map with a fixed cutting plane based on the vessel's draft. An example of such a map for a 7-meter draft vessel is displayed in Fig.~\ref{fig:islands}, which was generated using interpolation techniques on data collected by a series of depth sensors located along the river. It is worth noting that the size of the tide island varies over time for a given location in the waterway. Additionally, different sources of uncertainty, such as measurement noise or the effect of the moon, make it difficult to determine the exact shape of the island and the depth of the waterway at a specific location, thereby increasing the risk of grounding accidents.

Some works have paid particular attention to solving the scheduling problem in waterways affected by the tide, which complicates the problem due to the time-dependent depth of the waterway~\cite{li2021vessel}. However, despite all the efforts made to schedule and plan vessel trips in inland waterways influenced by tides, autonomous motion control of vessels in these channels remains a challenge, and expert river pilots are usually required to execute the optimal trip plans. Moreover, obtaining accurate dynamic models of time-varying depth for a specific waterway is a difficult task, if not impossible, due to their complexity. Meanwhile, interpolated models such as the one in Fig.~\ref{fig:islands} obtained from sensor readings are often used, which may not be accurate enough, jeopardizing security and increasing the risk of grounding accidents~\cite{ozlem2020grounding}.

\subsection{Our Contributions}

In the proposed framework, the vessel control problem is viewed as a trajectory-tracking problem, where the optimal reference trajectory is provided by a planner (e.g.,~\cite{nadales2023safe}). Due to the dynamic and uncertain nature of tide islands, it is crucial to design a motion controller that efficiently follows the reference trajectory while avoiding grounding accidents, especially when operating with a limited dataset of sensor readings.

Since the main source of uncertainty is the depth of the waterway at each point and time instance, we first propose a novel formulation to model tide islands as stochastic uni-dimensional polytopes in the time-space map where the constraint imposed by the depth of the waterway is violated. The only information available about these uncertainties is the limited set of data obtained from noisy measurements taken by depth sensors placed along the waterway. Using this dataset, an interpolated map of the time-space regions of depth constraint violation is obtained. Based on this information, a time-varying empirical distribution of the width of each tide island is derived. 

Then, we design a motion controller that assesses the risk of accidents via the conditional value-at-risk (CVaR), which measures the average worst-case loss. CVaR is a coherent risk measure in the sense of Artzner~\cite{artzner1999coherent} that can distinguish tail events and is considered an adequate risk measure for robotics applications~\cite{majumdar2020should}. However, in order to properly assess the risk of accidents, it is required to have accurate information about the probability distribution of the underlying uncertainties. Given that we have only a limited dataset of sensor readings, we may end up with undesired behaviors and a lack of safety. To address this issue, we utilize the distributionally robust risk-aware model predictive control (DR-MPC) method introduced in~\cite{hakobyan2021wasserstein}, which is robust against distributional inaccuracies within a specific set of distributions called the \emph{ambiguity set} and can result in high-performance and safe control actions even with inaccurate information. To the best of our knowledge, this marks the pioneering application of distributionally robust optimization (DRO) and distributionally robust control (DRC) in the field of maritime transportation.

In this work, the ambiguity set is designed as a statistical ball around the given nominal distribution, where the distance between distributions is measured via the Wasserstein distance. Unfortunately, the proposed DR-MPC problem is intractable as it constitutes an infinite-dimensional problem. To overcome this issue, a Kantorovich duality-based strategy~\cite{mohajerin2018data} is employed to reformulate the original problem into a finite-dimensional nonconvex optimization problem that can be solved using a standard commercial solver.

Finally, using depth sensor data collected in the Guadalquivir River, we empirically demonstrate the performance of our risk-aware Wasserstein DR-MPC approach that solves the motion control problem of vessels in the presence of inaccuracies in the empirical distribution. Our findings indicate that the approach successfully controls the vessel along the waterway without any collisions, even with a limited dataset of only six observations, while the standard non-robust, risk-aware model predictive controllers (MPC) result in collision rates of at least 63\%.\footnote{We define a collision as a situation in which a vessel enters an unsafe region due to a specific uncertainty realization. This definition serves as a measure of the algorithm's effectiveness in managing and ensuring that vessels remain within the unforeseen safe region, considering only a limited dataset of measurements.}

The remainder of this paper is organized as follows. In Section~\ref{sec:prelim}, we detail the preliminaries needed for solving the problem. Section~\ref{sec:wdro} presents the adapted version of the Wasserstein DR-MPC approach, explaining how tide islands are modeled as stochastic uni-dimensional polytopes and how the DR-MPC problem is formulated. In Section~\ref{sec:cs}, we apply and test the proposed motion control strategy in a real-world scenario in the Guadalquivir River, comparing the results to those obtained when using the empirical distribution directly. Finally, in Section~\ref{sec:conclusions}, we summarize the main conclusions of our work.

\section{Preliminaries}\label{sec:prelim}

\subsection{System and Obstacles Modeling}\label{subsec:obstacles}

In this work, we treat each vessel as a water surface robot whose dynamics can be modeled by a discrete-time system
\begin{equation*}
    \begin{aligned}
     x(t+1)& = f(x(t), u(t))\\
     y(t) &= h(x(t),u(t)),
    \end{aligned}
\end{equation*}
where $x(t)\in \mathbb{R}^{n_x}$ is the state of the system (e.g., location, velocity), and $u(t)\in \mathbb{R}^{n_u}$ is the control input (e.g., thrust, acceleration). The system output $y(t)\in \mathbb{R}^{n_y}$ represents the position of the vessel in a $n_y$-dimensional space. The functions $f:\mathbb{R}^{n_x}\times \mathbb{R}^{n_u}\rightarrow \mathbb{R}^{n_x}$ and $h:\mathbb{R}^{n_x}\times \mathbb{R}^{n_u}\rightarrow \mathbb{R}^{n_y}$ are general nonlinear functions representing the system dynamics and the output mapping, respectively. Perfect knowledge of the state and noiseless output measuring is assumed. Moreover, the system state and the control input are constrained as
\begin{equation*}
    x(t) \in \mathcal{X}, \ u(t) \in \mathcal{U},
\end{equation*}
where $\mathcal{X} \subseteq \mathbb{R}^{n_x}$ and $\mathcal{U} \subseteq \mathbb{R}^{n_u}$ are convex compact sets, designed based on the physical constraints. 

\begin{figure}[t]
     \centering
     \includegraphics[width=0.6\linewidth]{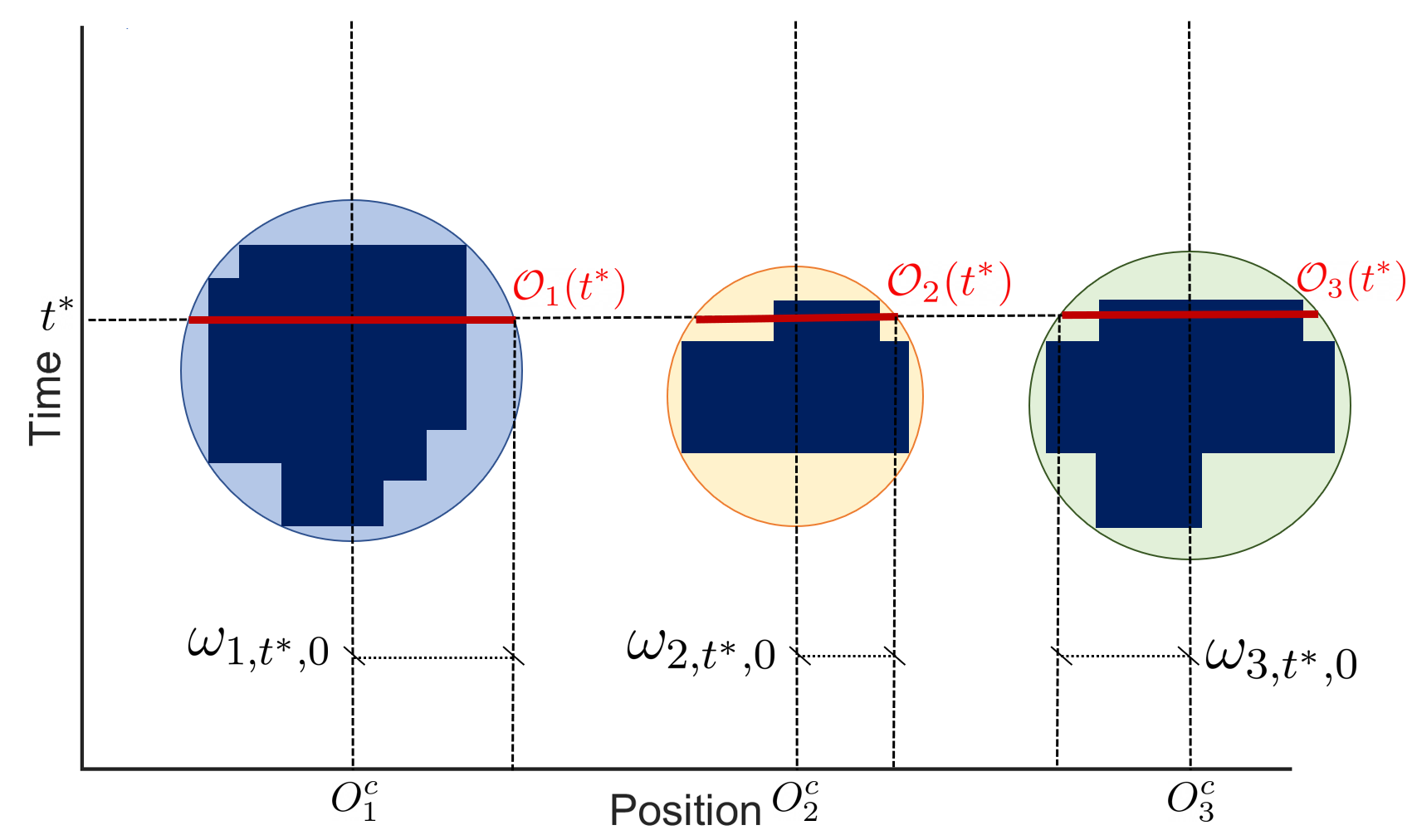}
 	\caption{Fictitious time-space map with random size-changing obstacles.}
 	\label{fig:obstacles}
\end{figure}

Vessels must safely sail through the waterway while avoiding obstacles,  which are the regions in the time-space map where the depth constraint is violated. Due to the nonconvex nature of these regions, each of them is approximated by the ball of minimum radius containing all points in the region, as it is depicted in Fig.~\ref{fig:obstacles}, where three fictitious tide islands have been represented\footnote{The approach proposed in this work is also applicable if the tide island is treated as the union of several convex polytopes. Instead, the approach of finding the minimum balls containing all points in the islands has been adopted for ease of explanation. Note that this approach results in a greater level of robustness.}. Here, the dark blue areas represent the tide islands, while $\omega_{l,t,k}$ represents the distance from the extreme of the ball to the center of each obstacle $l$ at time $t+k$.

Once the convex region containing each tide island is found, each of the obstacles to be avoided can be seen as a one-dimensional ball whose center is always located in the same position of the waterway but whose radii periodically change in time due to the effect of the tide. 
 In a general formulation of the collision avoidance problem, we consider a total of $L$ size-changing obstacles (tide islands) to be avoided by vessels while sailing in the waterway. The region occupied by each obstacle $l$, with $l=1,...,L$, at time instance $t$, is denoted by $\mathcal{O}_l(t) \subset \mathbb{R}$. The region occupied by each obstacle $l$ at time $t+k$ can be expressed as the following polytope:
 \begin{equation}\label{obs_def}
 \begin{split}
 \mathcal{O}_l (t+k) & := \{ x \in \mathbb{R} \mid | x - \mathcal{O}_l^c| \leq  \omega_{l,t,k}\}\\
 &= \{ x \in \mathbb{R} \mid A (x - \mathcal{O}_l^c) \leq  \omega_{l,t,k}\},
 \end{split}
 \end{equation}
where $O_l^c$ is the center of the ball, $\omega_{l,t,k}$ its radius at time $t+k$, while $A = [1, -1]^\top \in\mathbb{R}^{2}$. In this work, we consider $\omega_{l,t,k}$ to be a random variable whose true probability distribution is unknown. The obstacles' stochastic nature arises from various factors, including measurement noise from depth sensors, information loss during depth map interpolation, and the impact of unquantifiable natural phenomena on waterway depth (e.g., the effect of suspended sediments on the density of the river water and the tidal amplification effect~\cite{wang2014interaction}). 

The procedure of obtaining an empirical distribution for $\omega_{l,t,k}$ is detailed in Section~\ref{subsec:empirical}.  Note that these obstacles are given by the intersection of the ball containing all points where the depth constraint is violated with a plane of constant value given by the time instance considered. This is shown in Fig.~\ref{fig:obstacles}, where the red lines represent the obstacles to be avoided at some particular time $t^*$. 

Having the definition of each $l$-th obstacle, we define its safe region, $\mathcal{Y}_l (t+k)$, as the complement of the region occupied by the obstacle, i.e.,
\begin{equation*}
    \mathcal{Y}_l (t+k):= \mathbb{R}\setminus \mathcal{O}_l^\circ(t+k),
\end{equation*}
where $\mathcal{O}_l^\circ(t+k)$ represents the interior of $\mathcal{O}_l(t+k)$. For collision avoidance, vessels should always sail through the intersection of the safe regions associated with all obstacles, i.e.,
\[
y(t+k) \in \bigcap_{i=1}^{L}\mathcal{Y}_l (t+k),
\]
in order to guarantee safe navigation.

\begin{figure}[t]
\centering
\includegraphics[width=0.6\linewidth]{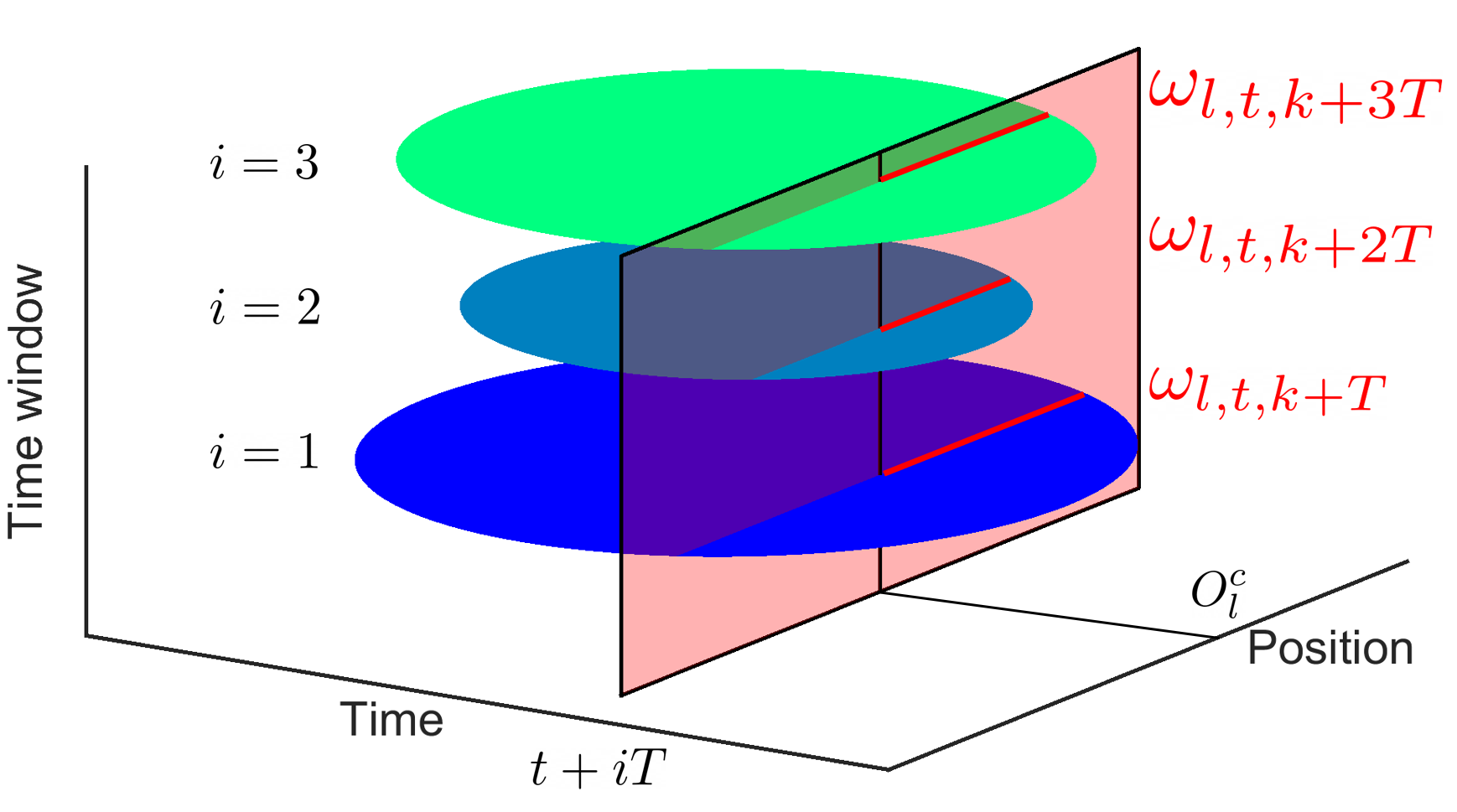}
\caption{Obtaining the uncertainty data set for $\omega_{l,t,k}$ at $t$ for different observations of the same obstacle  $l$.}
\label{fig:data_set}
\end{figure}

\subsection{Empirical Distribution Generation}\label{subsec:empirical}

In practice, the true distribution $\mu_{l,t,k}$ of the uncertain parameter $\omega_{l,t,k}$, representing the width of each obstacle $l$ at time $t+k$, is unknown, and the only information available about its distribution is a limited set of historical data. In this work, we assume that the tidal wave is a perfect sine of constant period. Therefore, we assume that the depth at time instances separated by intervals given by the tide period are samples drawn from the same probability distribution.

\begin{assumption}\label{assum:1}
Let the depth at some location of the waterway be a periodic wave~\cite{navarro2011temporal}. Let the period of this wave be given by a constant value, which is reasonable for the considered time period, given the small variation caused by the gravitational effect of the moon. Also, let  $T$ be the period of the tidal wave. Then the following expression holds:
\begin{equation}
    \omega_{l,t,k} := \omega_{l,t,k+iT}, \; i =1,...,N.\nonumber
\end{equation}
\end{assumption}

Under Assumption~\ref{assum:1}, a series of sets of all possible values of the width of each obstacle $l$ at each time instance $t$ has been obtained using the map of depth depicted in Fig.~\ref{fig:islands}. Each of these datasets is a collection of different observations of $\omega_{l,t,iT} $ for all $i = 1,...,N$, which is represented in Fig.~\ref{fig:data_set}. Then,  these datasets are employed to obtain an empirical distribution of $\omega_{l,t,k}$ for each obstacle and for each time instance.

To obtain an estimation of the real probability distribution, the only information available is a limited set of $N$ possibly noisy data samples, $\mathfrak{D}_{l,t,k} = \{\hat{\omega}_{l,t,k}^1,...,\hat{\omega}_{l,t,k}^N\}$. A possible option would be to obtain an empirical distribution defined as
\begin{equation*}
    \nu_{l,t,k} = \frac{1}{N}\sum_{i=1}^{N} \delta_{\hat{\omega}_{l,t,k}^i},
\end{equation*}
where $\delta_{\omega}$ is the Dirac delta function concentrated at $\omega$. This way of obtaining an empirical distribution is, however, not capable of providing a reliable estimation of the safety risk, especially when the size of the sample data set $N$ is small.

\section{Wassertein Distributionally Robust Motion Control}\label{sec:wdro}

Taking into account the changing depth and all operational and regulatory constraints for preventing collisions at sea~\cite{international2002colreg}, a reference trajectory can be obtained, for example, following the novel approach recently presented in~\cite{nadales2023safe}. Once the optimal reference trajectory for a vessel is determined, we adopt the Wasserstein DR-MPC control approach~\cite{hakobyan2021wasserstein} and apply it to track this trajectory while guiding the vessel to its final destination safely.
This approach limits the risk of grounding accidents caused by the waterway's shifting depths, which are treated as uncertainties subject to an unknown probability distribution.\footnote{Note that no assumptions are made about the distribution followed by each of the uncertainties affecting the size of the tidal islands. Our primary focus lies on capturing the overall robustness of the problem rather than explicitly characterizing the noise distribution. Adopting a DRC strategy allows us to account not only for the lack of information due to the limited size of the set of historical data but also for the sources of uncertainty affecting the measurements without explicitly identifying them.}

\subsection{Measuring the Risk of Unsafety Using CVaR}\label{subsec:safety}

The risk-aware control approach proposed in this work employs the notion of safety risk introduced in~\cite{samuelson2018safety}. In particular,  the loss of safety regarding obstacle $l$ be the distance between the vessel's location at time $t+l$ and its associated safe region, i.e., the vessel's penetration into the obstacle measured as
\begin{equation}\label{equ_dist}
    \dist(y(t+k), \mathcal{Y}_l(t+k)) := \min_{a \in \mathcal{Y}_l(t+k)} \lVert y(t+k) - a \rVert_2,
\end{equation}
where $\lVert \cdot \rVert_2$ is the standard Euclidean norm. The objective is to drive the vessel so that the loss of safety is zero. Otherwise, if this quantity takes a positive value, the vessel would penetrate into the obstacle, and a grounding accident would occur. However, the resulting actions will be overly conservative if a deterministic constraint is imposed on the penetration distance. As an alternative, we propose quantifying the unsafety associated with the obstacle $l$ using a risk measure. In particular, we measure the risk of unsafety by computing the CVaR of the loss of safety defined in~\eqref{equ_dist}.

The CVaR of a random loss $X$ following a distribution $\gamma$  is defined as the conditional expectation of the loss within the $(1-\alpha)$ worst-case quantile. As defined in \cite{rockafellar2002conditional}, it can be expressed as
\begin{equation*}
    \text{CVaR}_\alpha^\gamma (X) := \min_{z \in \mathbb{R}} \mathbb{E}^\gamma \bigg[ z + \frac{(X-z)^+}{1-\alpha}\bigg],
\end{equation*}
where $(\mathfrak{x})^+ = \max\{\mathfrak{x},0\}$. 
Thus, the safety risk defined by
\begin{equation*}
    \text{CVaR}_{\alpha}^{\mu_{l,t,k}}[\text{dist}(y(t+k),\mathcal{Y}_l(t+k))]
\end{equation*}
measures the conditional expectation of the distance between the vessel's position $y(t+k)$ and the safe region $\mathcal{Y}_l(t+k)$ within the $(1-\alpha)$ worst-case quantile of the safety loss distribution. Then, the risk constraint imposed on each vessel can be written as:
\begin{equation*}
    \text{CVaR}_{\alpha}^{\mu_{l,t,k}}[\text{dist}(y(t+k),\mathcal{Y}_l(t+k))] \leq \delta_l \ \forall l,
\end{equation*}
where $\delta_l \geq 0$ is a user-defined parameter that can be employed to adjust the risk tolerance of the vessel. It can be chosen to adjust the risk tolerance of the vessel, usually chosen to obtain the desired trade-off between robustness and optimal performance in terms of sailing times.

\subsection{Motion Control via DR-MPC}\label{subsec:drc}

The limitation imposed by the lack of information may lead to unsafe decision-making without respecting the original risk constraint, which could lead to a grounding accident as the safety risk is poorly assessed. To address this issue, the approach  proposed in~\cite{hakobyan2021wasserstein} is adopted, where the safety risk is evaluated under the worst-case distribution of $\omega_{l,t,k}$ belonging to a given set of distributions $\mathcal{D}_{l,t,k}$ called the \emph{ambiguity set}, leading to the following distributionally robust risk constraint:
\begin{equation}
    \sup_{\mu_{l,t,k} \in \mathcal{D}_{l,t,k}} \text{CVaR}_{\alpha}^{\mu_{l,t,k}}  [\dist(y(t+k), \mathcal{Y}_{l}(t+k)]\leq \delta_l \ \forall l. \label{eq:riskcon}
\end{equation}

By upper-bounding the worst-case risk value that the vessel can bear, we ensure that the resulting control action is robust against distribution errors. The construction of the ambiguity set is a crucial factor that impacts the trade-off between robustness and performance. Similar to~\cite{hakobyan2021wasserstein}, we model the ambiguity set as a statistical ball, which includes all probability measures whose distance from the empirical one is no greater than the given radius $\theta > 0$, i.e.,
\[
    \mathcal{D}_{l,t,k} : = \{\mu\in\mathcal{P}(\mathcal{W})\mid W_p(\mu, \nu_{l,t,k})\leq \theta \},
\]
where $\mathcal{P}(\mathcal{W})$ denotes the set of Borel probability measures on the support $\mathcal{W}\subseteq \mathbb{R}$. The distance between two measures is computed via the Wasserstein metric of order $p$~\cite{villani2009wasserstein}, which measures the minimum amount of work for transporting a certain mass from one distribution to another through non-uniform perturbations.

\begin{definition}
The $p$-order Wasserstein metric between two probability measures $\mu \in \mathcal{P}(\mathcal{W})$ and $\nu  \in \mathcal{P}(\mathcal{W})$ is defined as
\begin{equation*}
    W_p(\mu, \nu) : = \min_{\kappa \in \mathcal{P}(\mathcal{W}^2)}\left\{ \bigg(\int_{\mathcal{W}^2} \lVert \omega - \omega' \rVert^p  d\kappa(\omega, \omega')\bigg)^{1/p} \mid \Pi^1 \kappa = \nu , \Pi^2 \kappa = \mu \right\},
\end{equation*}
where $\kappa$ is the transportation plan, with $\Pi^i\kappa$ being its $i$th marginal, with $\lVert\cdot\rVert$ representing an arbitrary norm on $\mathbb{R}^n$.
\end{definition}

Though the Wasserstein metric has often been viewed in the context of the optimal transport problem~\cite{villani2021topics}, recent works in DRO and DRC have demonstrated the application and utility of Wasserstein ambiguity sets for promoting robustness in the distributional sense while providing substantial performance guarantees. Therefore, in this work, we aim to develop a control framework that is robust against distributional errors characterized by the Wasserstein ambiguity set.

\begin{remark}\label{remark:wassertein}
Other types of ambiguity sets can be employed in the DR-MPC formulation. A popular approach in the literature of DRO is the use of moment-based ambiguity sets~\cite{wiesemann2014distributionally}, which are often overly conservative, requiring a large amount of data to reliably estimate moment information. Another approach involves the use of statistical-distance-based ambiguity sets employing $\phi$-divergence~\cite{ben2013robust} and total variation distance~\cite{jiang2018risk}.
In contrast to these methods, Wasserstein ambiguity sets~\cite{gao2022distributionally} contain a richer set of relevant probability distributions. The associated DRO problem has been proven to offer superior finite-sample performance guarantees and address the closeness between two points in the support~\cite{mohajerin2018data}, especially useful in sequential decision-making problems~\cite{yang2020wasserstein}. Moreover, as demonstrated in~\cite{kuhn2019wasserstein}, Wasserstein DRO can alleviate overfitting due to its strong relation to regularization techniques used in machine learning algorithms. All these features make ambiguity sets based on Wasserstein distance an attractive approach for the proposed motion control tool.
\end{remark}

Taking this into account, the discrete-time Wasserstein DR-MPC problem for trajectory tracking can be formulated as
\begin{subequations}\label{mpc_problem}
\begin{align}
 \inf_{\mathbf{u},\mathbf{x},\mathbf{y}} \; & \lVert y_K-\tau_K\rVert_P^2+\sum_{k=0}^{K-1} \lVert y_k -\tau_k \rVert^2_Q + \sum_{k=1}^{K-1} \lVert \Delta u_k \rVert^2_R\label{cost:mpc_1}\\
 \st \; &x_{k+1} = f(x_k,u_k)  \label{con:mpc_1}\\
 & y_k = h(x_k,u_k) \label{con:mpc_2}\\
 & x_0 = x(t) \label{con:mpc_3}\\
 & x_k \in \mathcal{X} \label{con:mpc_4}\\
 & u_k \in \mathcal{U} \label{con:mpc_5}\\
 &\sup_{\mu_{l,t,k} \in \mathcal{D}_{l,t,k}} \text{CVaR}_{\alpha}^{\mu_{l,t,k}}  [\dist(y_k, \mathcal{Y}_{l}(t+k)] \leq \delta_l,  \label{con:mpc_6} 
\end{align}
\end{subequations}
where all constraints hold for $k=1,...,K$ and $l = 1,...,L$, except for~\eqref{con:mpc_1} and~\eqref{con:mpc_5} which hold for $k = 0,...,K-1$. Here, $K$ is the prediction horizon and  $\mathbf{u}:=(u_0,...,u_{K-1})$, $\mathbf{x}:=(x_0,...,x_{K})$, and $\mathbf{y}:=(y_0,...,y_{K})$, are the control input, state, and output sequences over the time horizon considered, respectively. The equality constraints~\eqref{con:mpc_1} and~\eqref{con:mpc_2} represent the system dynamics and the measurement of the predicted output, respectively. Polytopic constraints on state and control input are imposed via~\eqref{con:mpc_4} and \eqref{con:mpc_5}, while the initial state is set via~\eqref{con:mpc_3}. Finally,~\eqref{con:mpc_6} bounds the distributionally robust risk at a given tolerance level. The cost function~\eqref{cost:mpc_1} is chosen to penalize the deviation from the reference trajectory $\tau_k$ with weighting matrices $Q\succeq 0$ and $P\succeq 0$, as well as penalize large deviations of the control action $\Delta u_k := u_k - u_{k-1}$, which is included with a weighting matrix $R\succ 0$.
The Wasserstein DR-MPC is solved in a receding horizon fashion, where only the first component $u_0^*$  of the optimal solution $\mathbf{u}^*$ is applied to the system at time $t$, i.e., $u(t) = u_0^*$. However, solving this DR-MPC control problem is a challenging task since it involves an infinite-dimensional optimization problem over the space of distributions due to the risk constraint~\eqref{con:mpc_6}. To alleviate the intractability, we perform reformulation, detailed in the following subsection.

\begin{remark}\label{remark:alternatives}
In the proposed DR-MPC problem formulation,  we consider uncertainty in the obstacles' dimensions. This differs from~\cite{hakobyan2021wasserstein} in that the latter considers uncertainty in the obstacle's motion. The proposed formulation is also different from the one studied in~\cite{coulson2019regularized}, where authors consider uncertainties in the system. Several DRC problems solved via dynamic programming approaches can be found in the literature, as is the case of the problems studied in \cite{yang2017convex, yang2018dynamic, tzortzis2019infinite}, among others.
\end{remark}

\subsection{Finite-dimensional Reformulation via the Kantorovich Duality}\label{subsec:kantorovich}

In this subsection, we reformulate the DR-MPC problem to tackle the issue of infinite dimensionality and develop a tractable algorithm for solving it. For ease of explanation, the obstacle index and time subscripts will be omitted.  First, we derive a simple expression for the loss of safety \eqref{equ_dist}. Given that the region occupied by an obstacle $\mathcal{O}$ is a convex polytope, the corresponding safe region can be represented by the union of half-spaces as
\[
\mathcal{Y}:= \bigcup_{j=1}^{m} \big\{y \mid A_j (y - \mathcal{O}^c) \geq \omega \big\}.
\]
Then, we use the following lemma to reformulate the loss of safety using the half-space definition of the safe region.

\begin{lemma}\label{lemma:1}
Consider the obstacle $\mathcal{O}$ defined in~\eqref{obs_def} and the associated safe region $\mathcal{Y}$. Then, the loss of safety can be expressed as
\begin{equation}\label{loss}
    \dist(y, \mathcal{Y}) = \left[\omega + \min_{j=1,2} \left\{ A_j (\mathcal{O}^c -  y)\right\}\right]^+.
\end{equation}
\end{lemma}
The proof of this lemma can be found in Appendix~\ref{app:lem1}.

Next, we find a conservative approximation of the distributionally robust risk constraint given by~\eqref{eq:riskcon}. Specifically, as shown in~\cite{hakobyan2021wasserstein}, the distributionally robust safety risk can be upper-bounded by
\begin{equation*}
 \inf_{z\in \mathbb{R}} z  + \frac{1}{1-\alpha}  \sup_{\mu \in \mathcal{D}} \mathbb{E}^\mu \bigg[\max \Big\{\omega + \min_{j=1,2} \{A_j (\mathcal{O}^c -  y)\} -z,-z,0\Big\}\bigg].
\end{equation*}

This new risk formulation is still difficult to evaluate since it involves a maximization problem over a continuous set of distributions. Fortunately, when the distance between two distributions is measured via the Wasserstein distance of order one, we can reformulate the risk constraint using the Kantorovich duality~\cite{mohajerin2018data}, as proposed in~\cite{hakobyan2021wasserstein}, to transform it into a finite-dimensional optimization problem as follows.

\begin{proposition}\label{propo:1}
Consider the Wasserstein distance of order one. Let the uncertainty set be the entire space $\mathcal{W} = \mathbb{R}$ and $\|\cdot\|$ represent the Euclidean norm. Then, the following equality holds:
\begin{equation}\label{sup}
\sup_{\mu \in \mathcal{D}} \mathbb{E}^\mu \bigg[\max \Big\{\omega + \min_{j=1,2} \{A_j (\mathcal{O}^c -  y)\} -z,-z,0\Big\}\bigg]
= \left\{\begin{array}{@{}l@{\;}l@{}}
\inf_{\lambda, s}\; & \lambda\theta+\frac{1}{N}\sum_{i=1}^{N}s_i\\
\st \; & \hat{\omega}^{(i)}
 + \rho^\top A (\mathcal{O}^c -  y)  \leq s_i + z\\
 & \rho^\top e_2 = 1\\
&s_i+z\geq 0\\
&s_i \geq 0\\
& \lambda \geq 1\\
&\rho \geq 0,
\end{array}\right.
\end{equation}
where all constraints hold for $i = 1,...,N$. Here, $e_m \in \mathbb{R}^m$ is a vector whose elements are all equal to one.
\end{proposition}
The proof of this proposition can be found in Appendix~\ref{app:prop1}.
It is important to note that although Proposition~\ref{propo:1} is rooted in~\cite[Proposition 1]{hakobyan2021wasserstein}, the key difference lies in the decoupling of disturbances and the obstacle's position due to the way disturbances influence the system. This decoupling allows for constraint reduction when reformulating the DR-MPC problem. Specifically, the original DR-MPC problem~\eqref{mpc_problem} can be reformulated as
\begin{subequations}\label{mpc_problem_ref}
\begin{align}    \inf_{\substack{\mathbf{u},\mathbf{x},\mathbf{y},\\\mathbf{z}, \lambda, s, \mathbf{\rho}}} 
     & \lVert y_K-\tau_K\rVert_P^2+\sum_{k=0}^{K-1} \lVert y_k -\tau_k \rVert^2_Q + \sum_{k=1}^{K-1} \lVert \Delta u_k \rVert^2_R \\
     \st\  & x_{k+1} = f(x_k,u_k) \\
     & y_k = h(x_k,u_k)\\
     & x_0 = x(t)\\
     & z_{l,k} + \frac{1}{1-\alpha}\bigg[\lambda_{l,k}\theta +\frac{1}{N}\sum_{i = 1}^{N}s_{l,k,i}\bigg]\leq \delta_l \\
     &\hat{\omega}_{l,t,k}^{(i)} + \rho_{l,k}^\top A (\mathcal{O}_l^c - y_k) \leq s_{l,k,i} + z_{l,k}  \label{con:bilinear}\\
     & \rho_{l,k}^\top e_2 = 1 \\
     & s_{l,k,i} + z_{l,k} \geq 0 \\
     & s_{l,k,i} \geq 0 \\
     & \lambda_{l,k} \geq 1 \\
     &\rho_{l,k}\geq 0 \\
     & x_k \in \mathcal{X}\label{const_x}\\
     & u_k \in \mathcal{U} \label{const_u}\\
     & z_{l,k} \in \mathbb{R},
\end{align}
\end{subequations}
where all constraints hold for $k=1,...,K$, $l = 1,...,L$ and $i = 1,...,N$, except for constraints~\eqref{const_u} and~\eqref{const_x}, which hold for $k = 0,...,K-1$ and $k = 0,...,K$, respectively. The reformulated above problem is a finite-dimensional problem. However, it is still nonconvex due to the bilinearity of constraint~\eqref{con:bilinear}. A locally optimal solution could be obtained by means of nonlinear programming algorithms such as interior point methods or sequential quadratic programming~\cite{biegler2010nonlinear}. If the model of the system dynamics and output equations are affine, relaxation techniques can be employed to find a globally optimal solution~\cite{liberti2008introduction}.

A notable advantage of the Wasserstein DR-MPC approach~\eqref{mpc_problem_ref} is that for a carefully designed radius $\theta$, it assures a probabilistic finite-sample performance guarantee, meaning that the safety risk constraint is satisfied with a probability no less than a certain threshold, even when evaluated under a set of new samples chosen independently of the training data~\cite{hakobyan2021wasserstein}.

\begin{remark}\label{remark:deviation}
The deviation of the vessel from its reference trajectory could be such that maneuvers among vessels may not occur in the expected section of the waterway according to the optimal trip plan. To solve this, a multi-agent control approach could be adopted. The proposed method can be used for centralized control of multiple vessels by considering a single (possibly large) MPC problem considering all vessels and their associated operational constraints.  This scenario is out of the scope of this work, as we aim to propose a general formulation of the DR-MPC problem for a single vessel in natural waterways.
\end{remark}

\section{Case Study}\label{sec:cs}

In this section, we carry out a series of simulations in which the proposed control approach is applied to a real case scenario in the Guadalquivir River. We use the DR-MPC method as a tracking controller for the vessel of interest with its corresponding reference trajectory. For training purposes, it is assumed that only a set of $N=11$ possibly noisy values of past records of $\omega_{l,t,k}$ is available.  A series of experiments are carried out to analyze the performance and robustness of the proposed approach, measuring the cost of the DR-MPC problem and the minimum distance between the location of the vessel $y(t)$ and the contour of the safe region $\mathcal{Y}$. In the first one, for a given value of $\delta_l$, the risk tolerance level, we analyze the performance and robustness of the system as a function of $\theta$, the radius of the Wasserstein ball defining the ambiguity set centered at the empirical distribution.  Then, for a given value of $\theta$, we analyze the performance and safety of the vessel as a function of the number of samples $N$ employed to generate the empirical distribution.

All the results are compared to two baselines: a standard chance-constrained MPC (CC-MPC)~\cite{castillo2020real} and a risk-aware MPC (SAA-MPC)~\cite{hakobyan2019risk}, where in both cases the obstacle avoidance constraint solely relies on the empirical distribution. We maintained the same probability level $\alpha$ as used in DR-MPC for the baselines. All the experiments have been solved in AMD RYZEN 4000 CPU @ 3GHz using Gurobi solver, which incorporates a spatial branch and bound algorithm for solving bilinear programs~\cite{kolodziej2013global}.

\begin{figure}[t]
\centering
\includegraphics[width=0.7\linewidth]{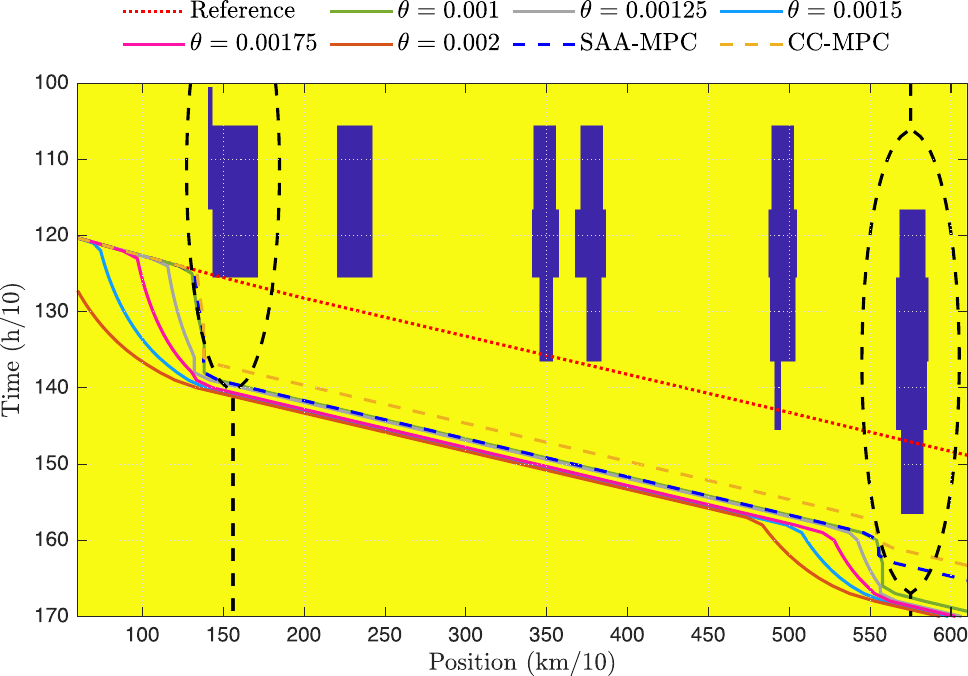}
\caption{Trajectories of the vessels controlled by the SAA-MPC, CC-MPC, and DR-MPC for different values of $\theta$. The dotted black lines around each tide island represent the maximum value of the uncertainty in the observation dataset.}
\label{fig:ex_1}
\end{figure}

\subsection{Setup}

The Guadalquivir River is divided into a total of 24 sections, representing the number of boundaries. Each of these sections $p$ is characterized by its length, minimum width, and the maximum speed at which vessels can sail through it. These parameter values can be found in~\cite{nadales2023safe}. The maximum time we allow each vessel to spend waiting and sailing through the waterway is set to 12. In the scheduling problem, a total of four vessels are considered. The maximum speed at which these vessels can sail is set to $v_i = 25 \, \mathrm{km/h}$, and the beam of each vessel is randomly selected. The draft of all vessels is set to $7 \, \mathrm{m}$. Thus, the depth map depicted in Fig.~\ref{fig:islands} is employed for all vessels, which shows that $L=5$ tide islands must be avoided by each vessel. The planning objective is then to minimize the time vessels spend sailing through the waterway and waiting to access it. We assume that the times when vessels arrive at the waterway are sampled from a Poisson distribution.

\subsection{Vessel Motion Control}

Given the optimal reference, we now demonstrate the utility of the proposed DRC method. The objective is to drive a single vessel from one extreme of the waterway to the other (i.e., track the reference trajectory) while avoiding the obstacles representing the tide islands.  Because the analysis of  vessel maneuverability is not the objective of this work, the model of the vessel is given by the linear system
\begin{equation*}
    x_{k+1} = x_k + T_s u_k,
\end{equation*}
where $x_k\in \mathbb{R}$ and $u_k\in \mathbb{R}$  are the position and speed of the vessel at time instance $k$, respectively, and $T_s$ is the sampling time, which in this case is set to $1$ h$/$10 (6 minutes).  Note that in this case, the measured state is the output of the system.
We set the MPC prediction horizon to $K=10$, while the weighting factors of the cost function are set to $Q=P=1$ and $R = 0.1$.

\begin{table}[t]
\centering
\caption{Average total cost, safety margin, and collision rate for the DR-MPC, SAA-MPC, and CC-MPC approaches for different values of $\theta$  with $\delta_l = 0.02$, $\alpha = 0.95$, and $N = 11$ computed over 100 simulations.}
\setlength{\tabcolsep}{0.5em} 
\renewcommand{\arraystretch}{1.2}
\begin{tabular}{>{\centering}m{1cm}| >{\raggedright}m{2.6cm} | >{\centering}m{3.5cm} | >{\centering}m{4cm} | >{\centering\arraybackslash}m{3.5cm}}
\hline
\rowcolor[HTML]{C0C0C0} 
\multicolumn{2}{c|}{\textbf{Method}} & \textbf{Total Cost} $(\times 10^{5})$ & \textbf{Safety Margin} (km) & \textbf{Collision Rate} $(\%)$ \\ \hline
\parbox[t]{2mm}{\multirow{5}{*}{\rotatebox[origin=c]{90}{\textbf{DR-MPC}}}} & $\theta=0.002$    & $3.495  \pm 0.068$          & $7.026 \pm 0.078$             & $0$         \\ \cline{2-5}
& $\theta=0.00175$  & $3.267 \pm 0.052 $          & $5.885 \pm 0.325$     & $0$                 \\ \cline{2-5}
& $\theta=0.0015$   & $3.079 \pm 0.064$          & $2.393  \pm 0.794$       & $0$             \\ \cline{2-5}
& $\theta=0.00125$  & $2.801 \pm 0.053$          & $1.041 \pm 0.587$       & $5$               \\ \cline{2-5}
& $\theta=0.001$    & $2.612 \pm 0.063$          & $-1.783 \pm 4.983$                &    $52$ \\ \hline
\multicolumn{2}{c|}{\textbf{SAA-MPC}} 
                  & $2.620 \pm 0.054$          & $-0.824 \pm 2.271$                     & $63$ \\ \hline
\multicolumn{2}{c|}{\textbf{CC-MPC}} 
                  & $1.739 \pm 0.062$          & $-21.801 \pm 0.808$                     & $89$ \\ \hline
\end{tabular}
 \label{table:ex1}
\end{table}

\subsubsection{Effect of Wasserstein Radius}

In the first experiment, we analyze the effect of the Wasserstein radius $\theta$ on the control performance and safety of the DR-MPC method. A total of $100$ simulations are performed for different values of $\theta$. We generate the empirical distribution assuming that, at each time step, only the $N$ last values of $\omega_{l,t,k}$ for each time step $t$ are available. These values are randomly selected from a validation data set of $1200$ values of $\omega_{l,t,k}$.

Examples of the different trajectories of the vessel controlled by the baselines and our DR-MPC with parameters $\delta_l = 0.02$, $\alpha = 0.95$, and $N = 11$ for a single uncertainty realization are shown in Fig.~\ref{fig:ex_1}. Note that, in this case, only the first and last obstacles affect the navigation due to their distances to the reference trajectory. For all the scenarios generated, the average value of the total costs, the minimum distance between the vessel and any of the obstacles (safety margin), as well as the collision rates are reported in Table~\ref{table:ex1}. Notably, larger values of $\theta$ correspond to higher costs for DR-MPC and larger safety margins. In contrast, both SAA-MPC and CC-MPC prioritize cost over safety, leading to negative safety margins and collisions occurring in 63\% and 89\% of cases, respectively. Remarkably, all trajectories generated by DR-MPC with $\theta \geq 0.0015$ are collision-free. Even for DR-MPC with $\theta=0.001$, the collision rate is lower than that for the baselines, underscoring the necessity of adopting the distributionally robust approach to ensure collision-free navigation. Moreover, our DR-MPC method achieves a positive average safety margin for $\theta \geq 0.00125$, indicating no collisions with obstacles. In contrast, both SAA-MPC and CC-MPC consistently show negative safety margins, signifying vessel penetration of obstacles.

\subsubsection{Effect of Sample Size}

In the second experiment, we examine how, for some fixed value of the Wasserstein radius, the control performance and safety are affected by the amount of data samples. It is assumed that only the last $N$ samples of $\omega_{l,t,k}$ for each time step $t$ are available. These values are randomly selected from a validation data set of $1200$ values of  $\omega_{l,t,k}$ for each time step $t$. The results of adopting the DR-MPC  approach are compared to those obtained when the baselines are applied for $100$ randomly generated scenarios using the same configuration as in the first experiment. Assuming the best trade-off between cost and safety is achieved at $\theta = 0.0015$, we present the average values of total cost, safety margin, and collision rate for DR-MPC, SAA-MPC, and CC-MPC in Table~\ref{table:ex2}. These results were obtained for different values of $N$, with $\delta_l = 0.02$ and $\alpha = 0.95$. As anticipated, trajectories generated by SAA-MPC and CC-MPC consistently result in collisions, regardless of the number of data samples considered. Moreover, the collision rate for vessels controlled by both SAA-MPC and CC-MPC rises as the number of samples decreases while the safety margin remains negative. In contrast, our DR-MPC method ensures collision-free navigation with a larger safety margin, even with a small dataset of $N=6$ samples.

\begin{table}[t!]
\centering
\caption{Average total cost, safety margin, and collision rate for DR-MPC, SAA-MPC, and CC-MPC approaches for different values of $N$  with $\delta_l = 0.02$, $\alpha = 0.95$, and $\theta = 0.0015$ computed over 100 simulations.}
\setlength{\tabcolsep}{0.5em} 
\renewcommand{\arraystretch}{1.2}
\begin{tabular}{>{\centering}m{1.6cm}| >{\raggedright}m{2.3cm} | >{\centering}m{3.3cm} | >{\centering}m{3.7cm} | >{\centering\arraybackslash}m{3.5cm}}
\hline
\rowcolor[HTML]{C0C0C0} 
\textbf{Method} & \textbf{Sample Size} & \textbf{Total Cost} $(\times 10^{5})$ & \textbf{Safety Margin} (km) & \textbf{Collision Rate} $(\%)$ \\ \hline
\parbox[t]{2mm}{\multirow{4}{*}{\rotatebox[origin=c]{90}{\textbf{DR-MPC}}}}  & $N=11$                          & $3.079 \pm 0.064$          & $2.393 \pm 0.794$          &       $0$         \\ \cline{2-5}
& $N=6$                          & $3.293 \pm 0.538$
& $1.272 \pm 0.963$          &       $0$         \\ \cline{2-5}
& $N=3$                           & $2.917 \pm 0.634$          & $ -6.436 \pm 8.712$        &    $34$              \\ \cline{2-5}
& $N=1$                           & $1.636 \pm 0.683$          & $ -18.673 \pm 10.095$        &     $100$             \\ \hline
\parbox[t]{2mm}{\multirow{4}{*}{\rotatebox[origin=c]{90}{\textbf{SAA-MPC}}}} & $N=11$                          & $2.620 \pm 0.054$          & $-0.824 \pm 2.271$          &       $63$         \\ \cline{2-5}
& $N=6$                          & $2.222 \pm 0.350$
& $-7.899 \pm 4.488$          &       $75$         \\ \cline{2-5}
& $N=3$                           & $1.963 \pm 0.424$          & $-16.729 \pm 8.605$        &    $100$              \\ \cline{2-5}
& $N=1$                           & $1.278 \pm 0.683$          & $-23.615 \pm 8.308$        &     $100$             \\ \hline
\parbox[t]{2mm}{\multirow{4}{*}{\rotatebox[origin=c]{90}{\textbf{CC-MPC}}}} & $N=11$    & $1.739 \pm 0.062$          & $-21.801 \pm 0.808$             & $89$         \\ \cline{2-5}
& $N=6$  & $1.783 \pm 0.261 $          & $-20.843 \pm 4.478$     & $96$                 \\ \cline{2-5}
& $N=3$   & $1.753 \pm 0.399$          & $-18.964 \pm 9.099$       & $100$             \\ \cline{2-5}
& $N=1$  & $1.268 \pm 0.591$          & $-22.805 \pm 9.060$       & $100$               \\ \hline
\end{tabular}
 \label{table:ex2}
\end{table}

\section{Conclusions}\label{sec:conclusions}

In this work, a DRC approach for vessels in natural inland waterways, paying particular attention to the case of the Guadalquivir River, where navigation is critically affected by the effect of the tide.  A risk-aware DR-MPC method was proposed for sailing through the waterway while avoiding stochastic grounding accidents. For that purpose, we modeled the regions of depth constraint violation as balls whose radii belong to an unknown probability distribution. We employ the dataset of past measurements of the tide to build the empirical distribution of the newly modeled obstacles' radii. To hedge against inaccuracies in the resulting distributional information, the DR-MPC method limits the safety risk by allowing perturbation within a Wasserstein ambiguity set. By doing so,  the proposed approach resolves the issue related to the inexact empirical distribution obtained from the limited amount of available data. After performing a series of experiments, the results obtained show that even with a very small sample size, the vessel driven by the Wasserstein DR-MPC successfully avoids the obstacles that randomly change in size over time. We also show how, for different values of the Wasserstein radius and the number of data samples, the proposed DR-MPC approach outperforms SAA-MPC and CC-MPC approaches, where only the empirical distribution is considered.

\appendix
\section{Proofs}
\subsection{Proof of Lemma~\ref{lemma:1}}\label{app:lem1}

\begin{proof}
Applying the steps in the proof of~\cite[Lemma~1]{hakobyan2021wasserstein}, it can be shown that
\[
\dist(y, \mathcal{Y}) 
= \left[\min_{j=1,2} \left\{\frac{\omega + A_j (\mathcal{O}^c - y)}{|A_j|}\right\}\right]^+.
\]
The proof is concluded using the fact that by definition $|A_j| = 1$ and taking $\omega$ outside the minimum.
\end{proof}

\subsection{Proof of Proposition~\ref{propo:1}}\label{app:prop1}
\begin{proof}
    Since the loss function~\eqref{loss} is affine in $\omega$, we first apply the results in~\cite{mohajerin2018data}[Corollary 5.1], so that the left-hand side of~\eqref{sup} equals to
    \begin{equation}\label{inf}
    \left\{\begin{array}{@{}l@{\;}l@{}}
    \inf_{\lambda, s}\; & \lambda\theta+\frac{1}{N}\sum_{i=1}^{N}s_i\\
    \st \; & \hat{\omega}^{(i)}
     + \min_{j=1,2} \{A_j (\mathcal{O}^c -  y)\} \leq s_i + z\\
    &s_i+z\geq 0\\
    &s_i \geq 0\\
    & \lambda \geq 1,
    \end{array}\right.
    \end{equation}
    where the constraints hold for $i=1,\dots,N$.  Next, we observe that
    \begin{equation}\label{fin_to_cont}
    \min_{j=1,2}\{A_j(\mathcal{O}^c - x)\} = \min_{\rho \geq 0}\{ \rho^\top A(\mathcal{O}^c - x) \mid \rho^\top e_2 = 1\}.
    \end{equation}
    Finally, it is required to show that the original problem~\eqref{sup} is equivalent to the reformulated problem~\eqref{inf}. Let $J^*$ and $J'$ be the optimal values of the original and reformulated problems, respectively. Let $(\lambda^*, s^*)$ and $(\lambda', s', \rho')$ denote the corresponding optimal solutions. By the feasibility of the point $s_i^*$ and by using~\eqref{fin_to_cont}, we have that
    \[
    \min_{j=1,2}\{A_j(\mathcal{O}^c - y)\} = \hat{\omega}^{(i)} + (\rho^*)^\top A(\mathcal{O}^c - y) \leq s_i^* + z,
    \]
    where $\rho^*$ is the optimal solution to the right-hand side in~\eqref{fin_to_cont}. Since all the other constraints are the same in both problems, we conclude that the optimal solution $(\lambda^*, s^*)$ to the original problem is feasible for the reformulation one, i.e., $J' \leq J^*$.

    To show that $J' \geq J^*$, we first see that
    \[
    \begin{split}
    \hat{\omega}^{(i)} + \min_{j=1,2}\{A_j(\mathcal{O}^c - y)\} &\leq \hat{\omega}^{(i)} + (\rho')^\top A(\mathcal{O}^c - y)\\
    &\leq s_i' + z,
    \end{split}
    \]
    where the inequalities follow from the feasibility of $\rho'$. By verifying all the remaining constraints, we conclude that the optimal solution $(\lambda', s', \rho')$ to the reformulated problem is feasible for the original one, i.e., $J' = J^*$. Thus, the two problems are equivalent.
\end{proof}

\bibliographystyle{IEEEtran}
\bibliography{bibliography.bib}

\begin{thebibliography}{10}
\providecommand{\url}[1]{#1}
\csname url@samestyle\endcsname
\providecommand{\newblock}{\relax}
\providecommand{\bibinfo}[2]{#2}
\providecommand{\BIBentrySTDinterwordspacing}{\spaceskip=0pt\relax}
\providecommand{\BIBentryALTinterwordstretchfactor}{4}
\providecommand{\BIBentryALTinterwordspacing}{\spaceskip=\fontdimen2\font plus
\BIBentryALTinterwordstretchfactor\fontdimen3\font minus \fontdimen4\font\relax}
\providecommand{\BIBforeignlanguage}[2]{{%
\expandafter\ifx\csname l@#1\endcsname\relax
\typeout{** WARNING: IEEEtran.bst: No hyphenation pattern has been}%
\typeout{** loaded for the language `#1'. Using the pattern for}%
\typeout{** the default language instead.}%
\else
\language=\csname l@#1\endcsname
\fi
#2}}
\providecommand{\BIBdecl}{\relax}
\BIBdecl

\bibitem{park2019role}
J.~S. Park, Y.-J. Seo, and M.-H. Ha, ``The role of maritime, land, and air transportation in economic growth: Panel evidence from {OECD} and non-{OECD} countries,'' \emph{Res. Transp. Econ.}, vol.~78, p. 100765, 2019.

\bibitem{li2022impact}
W.~Li, J.~Wu, H.~Du, Y.~Wan, S.~Yang, Y.~Xiao, and L.~Wang, ``Impact assessment of waterway development on the socioeconomic conditions and ecosystem in the upper {Y}angtze {R}iver,'' \emph{River Res. Appl.}, vol.~38, no.~5, pp. 988--999, 2022.

\bibitem{christodoulou2020forecasting}
A.~Christodoulou, P.~Christidis, and B.~Bisselink, ``Forecasting the impacts of climate change on inland waterways,'' \emph{Transp. Res. D: Transp. Environ.}, vol.~82, p. 102159, 2020.

\bibitem{kulkarni2020preventing}
K.~Kulkarni, F.~Goerlandt, J.~Li, O.~V. Banda, and P.~Kujala, ``Preventing shipping accidents: Past, present, and future of waterway risk management with {B}altic {S}ea focus,'' \emph{Saf. Sci.}, vol. 129, p. 104798, 2020.

\bibitem{costa2009establecimiento}
S.~Costa, J.~Guti{\'e}rrez~Mas, and J.~Morales, ``Establecimiento del r{\'e}gimen de flujo en el estuario del {G}uadalquivir, mediante el an{\'a}lisis de formas de fondo con sonda multihaz,'' \emph{Rev. Soc. Geol. Esp.}, vol.~22, no. 1-2, pp. 23--42, 2009.

\bibitem{losada2017tidal}
M.~Losada, M.~D{\'\i}ez-Minguito, and M.~Reyes-Merlo, ``Tidal-fluvial interaction in the {G}uadalquivir {R}iver {E}stuary: Spatial and frequency-dependent response of currents and water levels,'' \emph{J. Geophys. Res. Oceans}, vol. 122, no.~2, pp. 847--865, 2017.

\bibitem{donazar2018maintenance}
I.~Don{\'a}zar-Aramend{\'\i}a, J.~S{\'a}nchez-Moyano, I.~Garc{\'\i}a-Asencio, J.~Mir{\'o}, C.~Megina, and J.~Garc{\'\i}a-G{\'o}mez, ``Maintenance dredging impacts on a highly stressed estuary ({G}uadalquivir estuary): A {BACI} approach through oligohaline and polyhaline habitats,'' \emph{Mar. Environ. Res.}, vol. 140, pp. 455--467, 2018.

\bibitem{li2021vessel}
J.~Li, X.~Zhang, B.~Yang, and N.~Wang, ``Vessel traffic scheduling optimization for restricted channel in ports,'' \emph{Comput. Ind. Eng.}, vol. 152, p. 107014, 2021.

\bibitem{ozlem2020grounding}
{\c{S}}.~{\"O}zlem, Y.~C. Altan, E.~N. Otay, and {\.I}.~Or, ``Grounding probability in narrow waterways,'' \emph{J. Navig.}, vol.~73, no.~2, pp. 267--281, 2020.

\bibitem{nadales2023safe}
J.~M. Nadales, D.~Mu{\~n}oz de~la Pe{\~n}a, D.~Limon, and T.~Alamo, ``Safe optimal vessel planning on natural inland waterways,'' \emph{IEEE Trans. Intell. Transp. Syst.}, 2023.

\bibitem{artzner1999coherent}
P.~Artzner, F.~Delbaen, J.-M. Eber, and D.~Heath, ``Coherent measures of risk,'' \emph{Math. Finance}, vol.~9, no.~3, pp. 203--228, 1999.

\bibitem{majumdar2020should}
A.~Majumdar and M.~Pavone, ``How should a robot assess risk? towards an axiomatic theory of risk in robotics,'' in \emph{Robot. Res.}, 2020, pp. 75--84.

\bibitem{hakobyan2021wasserstein}
A.~Hakobyan and I.~Yang, ``Wasserstein distributionally robust motion control for collision avoidance using conditional value-at-risk,'' \emph{IEEE Trans. Robot.}, vol.~38, no.~2, pp. 939--957, 2021.

\bibitem{mohajerin2018data}
P.~Mohajerin~Esfahani and D.~Kuhn, ``Data-driven distributionally robust optimization using the {W}asserstein metric: Performance guarantees and tractable reformulations,'' \emph{Math. Program.}, vol. 171, no.~1, pp. 115--166, 2018.

\bibitem{wang2014interaction}
Z.~B. Wang, J.~C. Winterwerp, and Q.~He, ``Interaction between suspended sediment and tidal amplification in the {G}uadalquivir {E}stuary,'' \emph{Ocean Dyn.}, vol.~64, no.~10, pp. 1487--1498, 2014.

\bibitem{navarro2011temporal}
G.~Navarro, F.~J. Guti{\'e}rrez, M.~D{\'\i}ez-Minguito, M.~A. Losada, and J.~Ruiz, ``Temporal and spatial variability in the guadalquivir estuary: a challenge for real-time telemetry,'' \emph{Ocean Dyn.}, vol.~61, no.~6, pp. 753--765, 2011.

\bibitem{international2002colreg}
{I}nternational~{M}aritime {O}rganization, \emph{{COLREG}: {C}onvention on the {I}nternational {R}egulations for {P}reventing {C}ollisions at {S}ea, 1972}.\hskip 1em plus 0.5em minus 0.4em\relax International Maritime Organization, 2002.

\bibitem{samuelson2018safety}
S.~Samuelson and I.~Yang, ``Safety-aware optimal control of stochastic systems using conditional value-at-risk,'' in \emph{Proc. IEEE Am. Control Conf.}, 2018, pp. 6285--6290.

\bibitem{rockafellar2002conditional}
R.~T. Rockafellar and S.~Uryasev, ``Conditional value-at-risk for general loss distributions,'' \emph{J. Bank. Financ.}, vol.~26, no.~7, pp. 1443--1471, 2002.

\bibitem{villani2009wasserstein}
C.~Villani, ``The {W}asserstein distances,'' in \emph{Optim. Transp.}, 2009, pp. 93--111.

\bibitem{villani2021topics}
------, \emph{Topics in optimal transportation}.\hskip 1em plus 0.5em minus 0.4em\relax American Mathematical Soc., 2021, vol.~58.

\bibitem{wiesemann2014distributionally}
W.~Wiesemann, D.~Kuhn, and M.~Sim, ``Distributionally robust convex optimization,'' \emph{Oper. Res.}, vol.~62, no.~6, pp. 1358--1376, 2014.

\bibitem{ben2013robust}
A.~Ben-Tal, D.~Den~Hertog, A.~De~Waegenaere, B.~Melenberg, and G.~Rennen, ``Robust solutions of optimization problems affected by uncertain probabilities,'' \emph{Manage. Sci.}, vol.~59, no.~2, pp. 341--357, 2013.

\bibitem{jiang2018risk}
R.~Jiang and Y.~Guan, ``Risk-averse two-stage stochastic program with distributional ambiguity,'' \emph{Oper. Res.}, vol.~66, no.~5, pp. 1390--1405, 2018.

\bibitem{gao2022distributionally}
R.~Gao and A.~Kleywegt, ``Distributionally robust stochastic optimization with wasserstein distance,'' \emph{Math. Oper. Res.}, 2022.

\bibitem{yang2020wasserstein}
I.~Yang, ``Wasserstein distributionally robust stochastic control: A data-driven approach,'' \emph{IEEE Trans. Autom. Control}, vol.~66, no.~8, pp. 3863--3870, 2020.

\bibitem{kuhn2019wasserstein}
D.~Kuhn, P.~M. Esfahani, V.~A. Nguyen, and S.~Shafieezadeh-Abadeh, ``Wasserstein distributionally robust optimization: Theory and applications in machine learning,'' in \emph{Proc. Oper. Res. Manage. Sci. Age Anal.}, 2019, pp. 130--166.

\bibitem{coulson2019regularized}
J.~Coulson, J.~Lygeros, and F.~D{\"o}rfler, ``Regularized and distributionally robust data-enabled predictive control,'' in \emph{Proc. IEEE Conf. Decis. Control}, 2019, pp. 2696--2701.

\bibitem{yang2017convex}
I.~Yang, ``A convex optimization approach to distributionally robust {M}arkov decision processes with {W}asserstein distance,'' \emph{IEEE Contr. Syst. Lett.}, vol.~1, no.~1, pp. 164--169, 2017.

\bibitem{yang2018dynamic}
------, ``A dynamic game approach to distributionally robust safety specifications for stochastic systems,'' \emph{Automatica}, vol.~94, pp. 94--101, 2018.

\bibitem{tzortzis2019infinite}
I.~Tzortzis, C.~D. Charalambous, and T.~Charalambous, ``Infinite horizon average cost dynamic programming subject to total variation distance ambiguity,'' \emph{SIAM J. Control Optim.}, vol.~57, no.~4, pp. 2843--2872, 2019.

\bibitem{biegler2010nonlinear}
L.~T. Biegler, \emph{Nonlinear programming: concepts, algorithms, and applications to chemical processes}.\hskip 1em plus 0.5em minus 0.4em\relax SIAM, 2010.

\bibitem{liberti2008introduction}
L.~Liberti, ``Introduction to global optimization,'' \emph{Ecole Polytechnique}, 2008.

\bibitem{castillo2020real}
M.~Castillo-Lopez, P.~Ludivig, S.~A. Sajadi-Alamdari, J.~L. Sanchez-Lopez, M.~A. Olivares-Mendez, and H.~Voos, ``A real-time approach for chance-constrained motion planning with dynamic obstacles,'' \emph{IEEE Robot. and Autom. Lett.}, vol.~5, no.~2, pp. 3620--3625, 2020.

\bibitem{hakobyan2019risk}
A.~Hakobyan, G.~C. Kim, and I.~Yang, ``Risk-aware motion planning and control using {CVaR}-constrained optimization,'' \emph{IEEE Robot. Autom. Lett.}, vol.~4, no.~4, pp. 3924--3931, 2019.

\bibitem{kolodziej2013global}
S.~Kolodziej, P.~M. Castro, and I.~E. Grossmann, ``Global optimization of bilinear programs with a multiparametric disaggregation technique,'' \emph{J. Glob. Optim.}, vol.~57, no.~4, pp. 1039--1063, 2013.

\end{thebibliography}

\end{document}